%% file: main.tex
\pdfoutput=1

\documentclass[11pt]{article}

\usepackage[]{acl}

\usepackage{times}
\usepackage{latexsym}
\usepackage{algorithm}
\usepackage{algpseudocode}
\usepackage{fullwidth}

\usepackage[T1]{fontenc}

\usepackage[utf8]{inputenc}

\usepackage{microtype}
\usepackage{times}
\usepackage{helvet}
\usepackage{courier}
\usepackage{amsfonts,amsmath}
\usepackage{url}
\usepackage{paralist}
\usepackage{amsmath}
\usepackage{multirow}
\usepackage{multicol}
\usepackage{caption}
\usepackage{enumerate}
\usepackage{graphicx}
\usepackage{enumitem}
\usepackage{bm}
\usepackage[normalem]{ulem}
\usepackage{subfigure}
\usepackage{mathrsfs}
\usepackage{xcolor}
\usepackage{booktabs}

\newcommand{\cparagraph}[1]{\vspace{1.5mm}\noindent\textbf{#1}}

\begin{document}
%
\title{RecInDial: A Unified Framework for Conversational Recommendation \\ with Pretrained Language Models}
\author{Lingzhi Wang$^{1,2}$\thanks{\ \ Work performed during internship at Microsoft STCA.},
Huang Hu$^4$, Lei Sha$^3$, Can Xu$^4$, Kam-Fai Wong$^{1,2}$, Daxin Jiang$^4$\thanks{\ \ Corresponding author: djiang@microsoft.com.}\\
	$^1$The Chinese University of Hong Kong, Hong Kong, China \\
	$^2$MoE Key Laboratory of High Confidence Software Technologies, China \\
	$^3$University of Oxford, United Kingdom \\
	$^4$Microsoft Corporation, Beijing, China \\
    { \tt $^{1,2}$\{lzwang,kfwong\}@se.cuhk.edu.hk; $^3$lei.sha@cs.ox.ac.uk; } \\
    { \tt $^4$\{huahu,caxu,djiang\}@microsoft.com}
}
\maketitle
\input{sections/abstract}
\input{sections/introduction}

\input{sections/related_work}
\input{sections/model}

\input{sections/experimental_setup}
\input{sections/experimental_results}

\input{sections/conclusion}



\bibliography{anthology,custom}
\bibliographystyle{acl_natbib}

\end{document}

%% file: sections/abstract.tex
\begin{abstract}

Conversational Recommender System (CRS), which aims to recommend high-quality items to users through interactive conversations, has gained great research interest recently. A CRS is usually composed of a recommendation module and a generation module. In the previous work, these two modules are loosely connected in the model training and are shallowly integrated during inference, where a simple switching or copy mechanism is adopted to incorporate recommended items into generated responses. Moreover, the current end-to-end neural models trained on small crowd-sourcing datasets (e.g., 10K dialogs in the ReDial dataset) tend to overfit and have poor chit-chat ability. In this work, we propose a novel unified framework that integrates \underline{rec}ommendation \underline{in}to the \underline{dial}og (\textit{RecInDial}\footnote{The code is available at \url{https://github.com/Lingzhi-WANG/PLM-BasedCRS}}) generation by introducing a vocabulary pointer. To tackle the low-resource issue in CRS, we finetune the large-scale pretrained language models to generate fluent and diverse responses, and introduce a knowledge-aware bias learned from an entity-oriented knowledge graph to enhance the recommendation performance. Furthermore, we propose to evaluate the CRS models in an end-to-end manner, which can reflect the overall performance of the entire system rather than the performance of individual modules, compared to the separate evaluations of the two modules used in previous work. Experiments on the benchmark dataset ReDial show our RecInDial model significantly surpasses the state-of-the-art methods. More extensive analyses show the effectiveness of our model.

\end{abstract}

%% file: sections/introduction.tex
\section{Introduction}
\label{sec:intro}
In recent years, there have been fast-growing research interests to address Conversational Recommender System (CRS)~\cite{li2018towards,sun2018conversational,zhou2020improving}, due to the booming of intelligent agents in e-commerce platforms. 
It aims to recommend target items to users through interactive conversations. 
Traditional recommender systems perform personalized recommendations based on user's previous implicit feedback like clicking or purchasing histories, while CRS can proactively ask clarification questions and extract user preferences from conversation history to conduct precise recommendations. 
Existing generative methods~\cite{chen-etal-2019-towards,zhou2020improving,ma2020bridging,liang2021learning} are generally composed of two modules, \textit{i.e.}, a recommender module to predict precise items and a dialogue module to generate free-form natural responses containing the recommended items. Such methods usually utilize Copy Mechanism~\cite{gu2016incorporating} or Pointer Network~\cite{gulcehre2016pointing} to inject the recommended items into the generated replies. However, these strategies cannot always incorporate the recommended items into the generated responses precisely and appropriately. On the other hand, most of the existing CRS datasets~\cite{li2018towards,zhou2020towards,liu2020towards,liu2021durecdial} are relatively small ($\sim$10K dialogues) due to the expensive crowd-sourcing labor. 
The end-to-end neural models trained on these datasets from scratch are prone to be overfitting and have undesirable quality on the generated replies in practice. 

Encouraged by the compelling performance of pre-training techniques, we present a pre-trained language models (PLMs) based framework called \textit{RecInDial} to address these challenges.
\textit{RecInDial} integrates the item \underline{rec}ommendation \underline{in}to the \underline{dial}ogue generation under the pretrain-finetune schema.
Specifically, RecInDial finetunes the powerful PLMs like DialoGPT~\cite{zhang-etal-2020-dialogpt} together with a Relational Graph Convolutional Network (RGCN) to encode the node representation of an item-oriented knowledge graph. 
The former aims to generate fluent and diverse dialogue responses based on the strong language generation ability of PLMs, while the latter is to facilitate the item recommendation by learning better structural node representations.
To bridge the gap between response generation and item recommendation, we expand the generation vocabulary of PLMs to include an extra item vocabulary. 
Then a vocabulary pointer is introduced to control when to predict a target item from the item vocabulary or a word from the ordinary vocabulary in the generation process. 
The introduced item vocabulary and vocabulary pointer effectively unify the two individual processes of response generation and item recommendation into one single framework in a more consistent fashion.

\input{figures/intro_case}
To better illustrate the motivation of our work, Table \ref{tab:intro_case} shows a conversation example on looking for horrible movies and the corresponding replies generated by four models (\textit{ReDial} \cite{li2018towards}, \textit{KBRD} \cite{chen-etal-2019-towards}, \textit{KGSF} \cite{zhou2020improving}, OUR) together with the ground truth reply in the corpus (Human). As we can see, the previous work tends to generate short (e.g., ``KBRD: or It (2017)'') or in-coherent responses (e.g., ``KGSF: I would recommend watching it.''), which is resulted from the overfitting on the small dataset as we mentioned before. Different from them, our model can generate more informative and coherent sentences which shows a better chatting ability. In additon, 
we can notice that KGSF fails to raise a recommendation in the response ``I would recommend watching \underline{it}'' (``\underline{it}'' should be replaced with a specific item name in a successful combination of generation and recommendation results), which is probably due to the insufficient semantic knowledge learned and an ineffective copy mechanism. 
Our proposed unified PLM-based framework with a vocabulary pointer can effectively solve the issue.

Furthermore, to better investigate the end-to-end CRS system, we argue to evaluate the performance of recommendation by checking whether the final responses contain the target items. 
Existing works separately evaluate the performance of the two modules, \textit{i.e.}, dialogue generation and item recommendation. 
However, a copy mechanism or pointer network cannot always inject the recommended items into generated replies precisely and appropriately as we mentioned before. 
The performance of the final recommendations is actually lower than that of the recommender module.
For instance, the Recall@1 of the recommender module in KGSF~\cite{zhou2020improving} is 3.9\% while the actual performance is only 0.9\% when evaluating the final integrated responses (see Table~\ref{tab:main_rec}).

We conduct extensive experiments on the popular benchmark \textsc{ReDial}~\cite{li2018towards}. 
Our RecInDial model achieves a remarkable improvement on the recommendation over the state-of-the-art, and the generated responses are also significantly better on automatic metrics as well as human evaluation. 
Further ablation studies and quantitative and qualitative analyses demonstrate the superior performance of our approach. 

The contributions of this work can be: 
\begin{itemize}[leftmargin=*,topsep=2pt,itemsep=2pt,parsep=0pt]
    \item We propose a PLM-based framework called RecInDial for conversational recommendation. RecInDial finetunes the large-scale PLMs together with a Relational Graph Convolutional Network to address the low-resource challenge in the current CRS. 
    \item By introducing an extra item vocabulary with a vocabulary pointer, RecInDial effectively unifies two components of item recommendation and response generation into a PLM-based framework. 
    \item Extensive experiments show RecInDial significantly outperforms the state-of-the-art methods on the evaluation of both dialogue generation and recommendation.
\end{itemize}

%% file: figures/intro_case.tex
\begin{table}[t]\small
    \definecolor{mcolor1}{RGB}{41,18,171}
    \resizebox{\linewidth}{!}{\begin{tabular}{p{7.2cm}}
    \hline
    ... \\
    \hline
\textcolor{blue}{\textit{User}}: That sounds good. I could go with a classic. Have you seen \textcolor{mcolor1}{Troll 2 (1990)}? I'm looking for a horrible movie. cheesy horror \\
\hline
\textit{Human}: Tuesday 13, you like? \\
\textit{ReDial}: \textcolor{mcolor1}{Black Panther (2018)} is a good one too. \\
\textit{KBRD}: or \textcolor{mcolor1}{It  (2017)}\\
\textit{KGSF}: I would recommend watching it. \\
\textcolor{red}{\textit{OUR}}: yes I have seen that one. It was good. I also liked the movie \textcolor{mcolor1}{It (2017)}. \\
\hline
... \\
\hline
    \end{tabular}}
    \caption{A conversation example with \textcolor{mcolor1}{movies} recommendation from the test set of ReDial dataset.
    }
    
    \vskip -1em
    \label{tab:intro_case}
\end{table}

%% file: sections/related_work.tex
\section{Related Work}

Existing works in CRS can be mainly divided into two categories, namely attribute-based CRS and open-ended CRS.

\cparagraph{Attribute-based CRS.}
\indent
The attribute-based CRS can be viewed as a question-driven task-oriented dialogue system~\cite{zhang2018towards,sun2018conversational}.
This kind of system proactively asks clarification questions about the item attributes to infer user preferences, and thus search for the optimal candidates to recommend. 
There are various asking strategies studied by existing works, such as entropy-ranking based approach~\cite{wu2018q20}, generalized binary search based approaches~\cite{zou2019learning,zou2020towards}, reinforcement learning based approaches~\cite{chen2018learning,lei2020estimation,deng2021unified}, adversarial learning based approach~\cite{ren2020crsal} and graph based approaches~\cite{xu2020user,lei2020interactive,ren2021learning,xu2021adapting}.
Another line of research on this direction address the trade-off issue between exploration (\textit{i.e.}, asking questions) and exploitation (\textit{i.e.}, making recommendations) to achieve both the engaging conversations and successful recommendations, especially for the cold-start users. 
Some of them leverage bandit online recommendation methods to address cold-start scenarios~\cite{li2010contextual,li2016collaborative,christakopoulou2016towards,li2020seamlessly}, while others focus on the asking strategy with fewer turns~\cite{lei2020estimation,lei2020interactive,shi2019we,sun2018conversational}.

\cparagraph{Open-ended CRS.}
\indent
Existing works \cite{li2018towards,lei2018sequicity,jiang2019improving,ren2020thinking,hayati2020inspired,ma2020bridging,liu2020towards,wang2022improving} on this direction explore CRS through more free-form conversations, including proactively asking clarification questions, chatting with users, providing the recommendation, etc.
Multiple datasets have been released to help push forward the research in this area, such as \textsc{ReDial}~\cite{li2018towards},
\textsc{TG-Redial} (Chinese)~\cite{zhou2020towards}, \textsc{INSPIRED}~\cite{hayati2020inspired} and DuRecDial~\cite{liu2020towards,liu2021durecdial}. 
\citet{li2018towards} make the first attempt on this direction and contribute the benchmark dataset \textsc{ReDial} by the paired crowd-workers (\textit{i.e.}, Seeker and Recommender).
Follow-up studies~\cite{chen-etal-2019-towards,zhou2020improving,zhou2020towards} leverage the multiple external knowledge to enhance the performance of open-ended CRS.
CR-Walker~\cite{ma2020bridging} is proposed to perform the tree-structured reasoning on the knowledge graph to introduce relevant items, while MGCG~\cite{liu2020towards} addresses the transition policy from a non-recommendation dialogue to a recommendation-oriented one. 
Besides, \citet{zhou2021crslab} develop an open-source toolkit CRSLab to further facilitate the research on this direction. 
Most of these works utilize pointer network~\cite{gulcehre2016pointing} or copy mechanism~\cite{gu2016incorporating,sha2018order} to inject the recommended items into generated replies. 
Our work lies in the research of open-ended CRS.
While different from the previous work, we present a PLM-based framework for CRS, which finetunes the large-scale PLMs together with a pre-trained Relational Graph Convolutional Network (RGCN) to address the low-resource challenge in CRS. 

Another line of related work lies in the end-to-end task-oriented dialogs~\cite{wu2019global,he2020fg2seq,raghu2021constraint}, which also require response generation based on a knowledge base but not for recommendations.

%% file: sections/model.tex
\section{Methodology}
\input{figures/framework}
In this section, we present our proposed RecInDial model. Figure \ref{fig:sketch} shows the model overview. 
We first formalize the conversational recommendation task and then detail our PLM-based response generation module together with the vocabulary pointer. After that, we introduce how to incorporate the knowledge from an item-oriented knowledge graph with an RGCN into the model. 
Finally, we describe the model training objectives.

\subsection{Problem Formalization}
\label{subsec:problem_formulation}
The input of a CRS model contains the history context of a conversation, which is denoted as a sequence of utterances $\{t_1, t_2, ..., t_m\}$ in chronological order 
($m$ represents 
the number of utterances). 
Each utterance is either given by the seeker (user) or recommender (the model), which contains the token sequence $\{w_{i,1}, w_{i,2}, ..., w_{i,n_i}\}$ ($1\le i\le m$), where $w_{ij}$ is the $j$-th token in the $i$-th utterance and $n_i$ is the number of tokens in $i$-th utterance. 
Note that we define the name of an item as a single token and do not tokenize it. 
The output token sequence by the model is denoted as $\{w_{n+1}, w_{n+2}, ..., w_{n+k}\}$, where $k$ is the number of generated tokens and $n=\sum_1^{m}n_i$ is the total number of tokens in context. 
When the model conducts the recommendation, it will generate an item token $w_{n+i}$ ($1\le i\le k$) together with the corresponding context. 
In this way, recommendation item and response are generated concurrently. 

\subsection{Response Generation Model}
\label{subsec:generation}
In this subsection, we introduce how to extend PLMs to handle CRS task and produce items recommendation during the dialogue generation. 

\paragraph{PLM-based Response Generation.}

Given the input (\textit{i.e.}, the conversation history context $\{t_1, t_2, ..., t_m\}$), we concatenate the history utterances into the context $C = \{w_1, w_2, ..., w_n\}$ where $n$ is the total number of tokens in the context. 
Then the probability of the generated response $R = \{w_{n+1}, w_{n+2}, ..., w_{n+k}\}$ is formulated as:
\vskip -1.5em
\begin{equation}\small
\label{eq:gpt2}
\mathrm{PLM}(R|C) = \prod_{i=n+1}^{n+k} p(w_i|w_1,...,w_{i-1}).
\end{equation}
\vskip -0.5em
\noindent where $\mathrm{PLM}(\cdot|\cdot)$ denotes the PLMs of Transformer~\cite{vaswani2017attention} architecture. For a multi-turn conversation, we can construct $N$ such context-response pairs, where $N$ is the number of utterances by the recommender. 
Then we finetune the PLMs on all possible $(C,R)$ pairs constructed from the dialogue corpus. 
By this means, not only does our model inherit the strong language generation ability of the PLMs, but also simultaneously can learn how to generate the recommendation utterances on the relatively small CRS dataset.

\paragraph{PLM-based Item Generation.} 
To integrate the item recommendation into the generation process of PLMs, we propose to expand the generation vocabulary of PLMs by including an extra item vocabulary. We devise a vocabulary pointer to control when to generate tokens from the ordinary vocabulary or from the item vocabulary. 
Concretely, we regard an item as a single token and add all items into the item vocabulary. 
Hence, our model can learn the relationship between context words and candidate items. 
Such a process integrates the response generation and item recommendation into a unified model that can perform the end-to-end recommendation through dialogue generation.

\paragraph{Vocabulary Pointer.} 
We first preprocess the dialogue corpus and introduce two special tokens \texttt{[RecS]} and \texttt{[RecE]} to indicate the start and end positions of the item in utterance. 
Then we divide the whole vocabulary $V$ into $V_G$ and $V_R$, where $V_G$ includes the general tokens (\textit{i.e.}, tokens in the original vocabulary of PLM) and \texttt{[RecS]} while $V_R$ contains the all item tokens and \texttt{[RecE]}.
We then introduce a binary \textit{Vocabulary Pointer} $I_{vp}$ to guide the generation from $V_G$ or $V_R$. 
The model generates tokens in $V_G$ when $I_{vp}=0$, and generates the tokens in $V_R$ when $I_{vp}=1$, which can be formulated as follows:
\vskip -1.5em
\begin{equation}\small
\label{eq:vp1}
p(w=w_i) = \frac{exp(\phi_I(w_i)+\tilde{h}_i)}{\sum_{w_j \in V}exp(\phi_I(w_j)+\tilde{h}_j) }
\end{equation}
\vskip -1em
\begin{equation}\small
\label{eq:vp2}
\phi_I(w_j) = 
\begin{cases} 0, & I_{vp}=0, w_j \in V_G \,\text{or}\\ & I_{vp}=1, w_j \in V_R, \\
-inf, & I_{vp}=1, w_j \in V_G \,\text{or}\\ & I_{vp}=0, w_j \in V_R  \end{cases},
\end{equation}
\vskip -0.5em
\noindent where $\tilde{h} = h_L W_e^T$ is the feature vector before the softmax layer in Figure~\ref{fig:sketch}, $\tilde{h}_i$ means the feature value of the $i$-th token. $I_{vp}$ is initialized as $0$ at the beginning of the generation and won't change until the model produces \texttt{[RecS]} or \texttt{[RecE]}. 
It changes to $1$ if the model produces \texttt{[RecS]} (\textit{i.e.}, the model begins to generate items) and changes back to $0$ if \texttt{[RecE]} is emitted. 
Such a procedure continues until the turn is finished. 
With the \textit{Vocabulary Pointer}, our model can alternatively switch between generating response words and recommending items based on its previous outputs in a unified fashion. 
\input{tables/algorithm}

To help readers better understand the Vocabulary Pointer mechanism, we summarize the process in Algorithm \ref{alg:cap}.

\subsection{Knowledge Graph Enhanced Finetuning}
\label{subsec:vocab_bias}
Due to the difficulty of fully understanding user preferences by the conversation context, it is necessary to introduce the external knowledge to encode the user preferences when finetuning response generation model.
Inspired by the previous work \cite{chen-etal-2019-towards,zhou2020improving}, we also employ a knowledge graph from DBpedia \cite{lehmann2015dbpedia} and perform entity linking~\cite{daiber2013improving} to the items in the dataset, which helps better model the user preferences.
A triple in DBpedia is denoted by $<e_1, r, e_2>$, where $e_1, e_2 \in \mathcal{E}$ are items or entities from the entity set $\mathcal{E}$ and $r$ is entity relation from the relation set $\mathcal{R}$. 

\noindent\textbf{Relational Graph Propagation.}
We utilize R-GCN \cite{schlichtkrull2018modeling} to encode structural and relational information in the knowledge graph to entity hidden representations. 
Formally, the representation of node $e$ at $(l+1)$-th layer is: 
\vskip -1em
\begin{equation}\small
\bm{h}_e^{(l+1)} = \sigma (\sum_{r\in\mathcal{R}} \sum_{e^{\prime}\in\mathcal{E}_e^r} \frac{1}{Z_{e,r}}\bm{W}_r^{(l)}\bm{h}_{e^{\prime}}^{(l)} + \bm{W}^{(l)}\bm{h}_e^{(l)}),
\end{equation}
\vskip -0.5em
\noindent where $\bm{h}_e^{(l)} \in \mathbb{R}^{d_E}$ is the node representation of $e$ at the $l$-th layer, and $\mathcal{E}_e^r$ denotes the set of neighboring nodes for $e$ under the relation $r$. 
$\bm{W}_r^{(l)}$ is a learnable relation-specific transformation matrix for the embedding from neighboring nodes with relation $r$, while $\bm{W}^{(l)}$ is another learnable matrix for transforming the representations of nodes at the $l$-th layer and $Z_{e,r}$ is a normalization factor. 

At the last layer $L$, structural and relational information is encoded into the entity representation $\bm{h}_e^{(L)}$ for each $e \in \mathcal{E}$. 
The resulting knowledge-enhanced hidden representation matrix for entities in $\mathcal{E}$ is denoted as $\bm{H}^{(L)} \in \mathbb{R}^{|\mathcal{E}| \times d_E}$. 
We omit the (L) in the following paragraphs for simplicity.

\noindent\textbf{Entity Attention.} 
Given a conversation context, we first collect the entities appeared in the context, and then we represent the user preference as $\mathcal{T}_u = {e_1, e_2, ..., e_{|\mathcal{T}_u|}}$, where $e_i \in \mathcal{E}$. 
After looking up the knowledge-enhanced representation table of entities in $\mathcal{T}_u$ from $\bm{H}$, we get:
\vskip -0.5em
\begin{equation}\small
\label{eq:entity}
\bm{H}_u = (\bm{h}_1, \bm{h}_2, ..., \bm{h}_{|\mathcal{T}_u|}),
\end{equation}
\noindent where $\bm{h}_i \in \mathbb{R}^{d_E}$ is the hidden vector of entity $e_i$. 
Then the self-attention mechanism \cite{lin2017structured} is applied to $\bm{H}_u$, which outputs a distribution $\alpha_u$ over $|\mathcal{T}_u|$ vectors:
\vskip -0.5em
\begin{equation}\small
\alpha_u = softmax(\bm{w}_{a2}tanh(\bm{W}_{a1}\bm{H}_u^T)),
\end{equation}
\noindent where $\bm{W}_{a1} \in \mathbb{R}^{d_a \times d_E}$ and $\bm{w}_{a2} \in \mathbb{R}^{1 \times d_a}$ are learnable parameters. Then we get the final representation for user history $u$ as follows: 
\vskip -1em
\begin{equation}\small
\label{eq:user_representation}
\bm{t}_u = \alpha_u\bm{H}_u.
\end{equation}
\vskip -0.5em
\noindent\textbf{Knowledge-Aware Bias.} 
To incorporate the knowledge from the constructed knowledge graph into our model while generating recommendation items, we first map the derived user representation $\bm{t}_u$ into the item vocabulary space $|V_R|$ as follows: 
\vskip -1em
\begin{equation}\small
\label{eq:bias}
\bm{b}_u = \bm{t}_u\bm{H}^T\bm{M}_b,
\end{equation}
\vskip -0.5em
\noindent where $\bm{M}_b \in \mathbb{R}^{|\mathcal{E}| \times |V_R|}$ are learnable parameters. 
Then we add $b_u$ to the projection outputs before softmax operation in the generation as a bias.
In this way, our model can produce items in aware of their relational knowledge and thus enhance the performance of recommendation.

\subsection{Recommendation in Beam Search}
\label{subsec:beam_search} 
To embed the top-k item recommendation into the generation, we develop a revised beam search decoding. 
Specifically, when we finish the generation for one response, we first check whether it contains the item names (i.e., whether it generates recommendations). 
If yes, then we choose the top-k items between \texttt{[RecS]} and \texttt{[RecE]} according to the probability scores at current time-step.

\subsection{Learning Objectives}
\label{subsec:model_training} 
There are two objectives, \textit{i.e.}, node representation learning on knowledge graph and the finetuning of response generation model.
For the former, we optimize the R-GCN and the self-attention network based on the cross entropy of item prediction: 
\vskip -1.5em
\begin{equation}\small
\mathcal{L}_{kg} = \sum_{(u, i) \in \mathcal{D}_1} -log(\frac{exp(\bm{t}_u\bm{H}^T)_i}{\sum_j exp(\bm{t}_u\bm{H}^T)_j}),
\end{equation}
\vskip -0.5em
\noindent where the item $i$ is the ground-truth item and $u$ is the corresponding user history, while $\mathcal{D}_1$ contains all training instances and $\bm{t}_u\bm{H}^T \in \mathbb{R}^{|\mathcal{E}|}$.

For the latter, we optimize another cross entropy loss for all generated responses, denoted as $R$. The following formula summarizes the process:
\vskip -1.5em
\begin{equation}\small
\mathcal{L}_{gen} = \sum_{(C,R) \in \mathcal{D}_2} \sum_{w_i \in R}-log(p(w_i|w_{<i},C)),
\end{equation}
\vskip -0.5em
\noindent where $p(w_i)$ refers to Eq.~\ref{eq:vp1} and $\mathcal{D}_2$ contains all $(C,R)$ pairs constructed from the dataset. We train the whole model end-to-end with the joint effects of the two objectives $\mathcal{L}_{kg}+\mathcal{L}_{gen}$.

%% file: figures/framework.tex
\begin{figure}[t]
\centering
\includegraphics[width=0.45\textwidth]{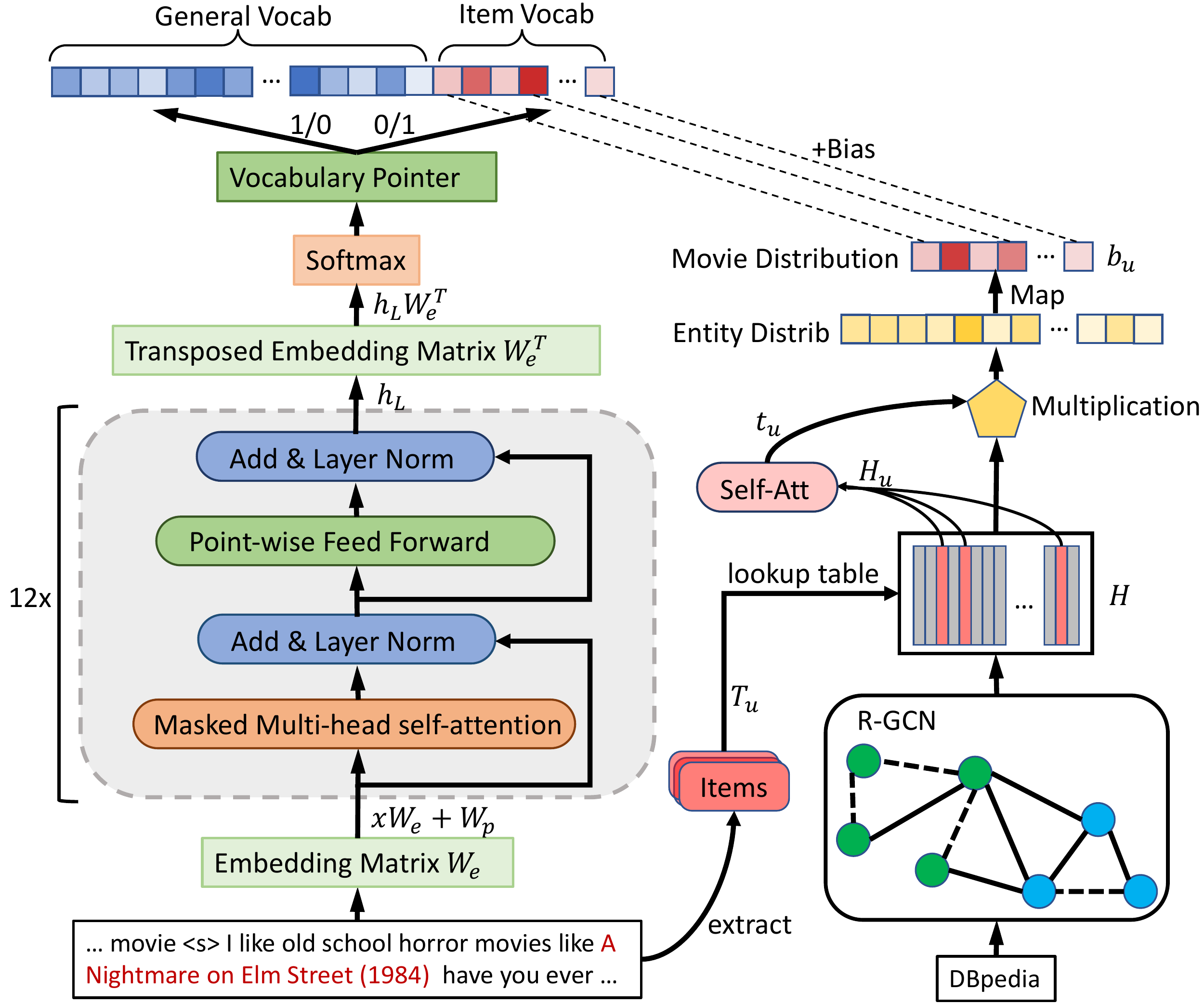}
\vskip -0.5em
\caption{\label{fig:sketch} Model overview of RecInDial.}
\vskip -1.5em
\end{figure}

%% file: tables/algorithm.tex
\begin{algorithm}[t]\small
\renewcommand{\algorithmicrequire}{\textbf{Input:}}
\renewcommand{\algorithmicensure}{\textbf{Output:}}
\caption{Vocabulary Pointer based Generation for RecInDial}\label{alg:cap}
\begin{algorithmic}
\Require history context $C$, general and item vocabulary $V_G$, $V_R$
\Ensure generated response $R$
\State extract appeared entities from $C$ as user preference $\mathcal{T}_u$
\State compute knowledge-aware bias $\bm{b}_u$ based on $\mathcal{T}_u$ using Eq.~\ref{eq:entity} to \ref{eq:bias}
\State $R \gets \{\}$
\State $n \gets 0$
\State $I_{vp} \gets 0$, $V \gets V_G$
\While{$n < N_{max}$} 
\State $w_n = Decode(C \bigcup R, V, \bm{b}_u)$
\textcolor[rgb]{0.6,0.6,0.6}{\Comment{Generate $w_n$ based on the previous tokens and bias from $V$}}
\State $R \gets R \bigcup\{w_n\}$
\If{$w_n = [RecS]$} \textcolor[rgb]{0.6,0.6,0.6}{\Comment{Generate tokens from $V_R$}}
\State $I_{vp} \gets 1$, $V \gets V_R$
\ElsIf{$w_n = [RecE]$}
\textcolor[rgb]{0.6,0.6,0.6}{\Comment{Generate tokens from $V_G$}}
\State $I_{vp} \gets 0$, $V \gets V_G$
\ElsIf{$w_n = [EOS]$}
\textcolor[rgb]{0.6,0.6,0.6}{\Comment{Generation is done}}
\State \textbf{break}
\EndIf
\State $n \gets n+1$
\EndWhile
\State \textbf{return} R
\end{algorithmic}
\end{algorithm}

%% file: sections/experimental_setup.tex
\section{Experimental Setup}
\label{sec:exp_setup}
\paragraph{Datasets.}
We evaluate our model on the benchmark dataset \textsc{ReDial}~\cite{li2018towards}. Due to the collection difficulty of the real world data, most the previous work \cite{li2018towards,chen-etal-2019-towards,zhou2020improving} only conducts experiments on this single dataset. 
The statistics of \textsc{ReDial} dataset is shown in Table~\ref{tab:dataset_statistic}. Detailed statistics of movie mentions are shown in Figure \ref{sfig:statistis_fre}. 
Most of the movies occur less than 5 times in the dataset, which indicates an obvious data imbalance problem in the \textsc{ReDial}. 
We also show the relationship between the average number of movie mentions and the number of dialog turns in Figure~\ref{sfig:statistic_turn}.
As we can see, there are less than 2 movie mentions when the dialogue turn number is less than 5. Finally, we follow \cite{li2018towards} to split the dataset into 80-10-10, for training, validation and test.

\paragraph{Parameter Setting.}
We finetune the small size pre-trained DialoGPT model\footnote{\url{https://huggingface.co/microsoft/DialoGPT-small}}, which consists of 12 transformer layers. The dimension of embeddings is 768. It is trained on 147M multi-turn dialogues from Reddit discussion threads.  
For the knowledge graph (KG), both the entity embedding size and the hidden representation size are set to 128, and we set the layer number for R-GCN to 1. 
For BART baseline, we finetune the base model \footnote{\url{https://huggingface.co/facebook/bart-base}} with 6 layers in each of the encoder and decoder, and a hidden size of 1024. For GPT-2 baseline, we finetune the small model\footnote{\url{https://huggingface.co/gpt2}}. 
For all model's training, we adopt Adam optimizer and the learning rate is chosen from \{$1e-5$, $1e-4$\}. The batch size is chosen from \{32, 64\}, the gradient accumulation step is set to 8, and the warm-up step is chosen from \{500, 800, 1000\}. All the hyper-parameters are determined by grid-search.
\input{tables/dataset_statistic}
\input{figures/data_statistic}

\paragraph{Baselines and Comparisons.}
We first introduce two baselines for recommender and dialogue modules, respectively. 
(1) \underline{\textbf{Popularity}}. It ranks the movie items according to their historical frequency in the training set without a dialogue module. 
(2) \underline{\textbf{Transformer}}~\cite{vaswani2017attention}. It utilizes a transformer-based encoder-decoder to generate responses without recommender module.

We then compare the following baseline models in the experiment:
(3) \underline{\textbf{ReDial}}~\cite{li2018towards}. It consists of a dialogue generation module based on HRED~\cite{serban2017hierarchical}, a recommender module based on auto-encoder~\cite{he2017distributed}, and a sentiment analysis module. 
(4) \underline{\textbf{KBRD}}~\cite{chen-etal-2019-towards}. It utilizes a knowledge graph from DBpedia to model the relational knowledge of contextual items or entities, and the dialogue generation module is based on the transformer architecture.
(5) \underline{\textbf{KGSF}}~\cite{zhou2020improving}. It incorporates and fuses both word-level and entity-level knowledge graphs to learn better semantic representations for user preferences. 
(6) \underline{\textbf{GPT-2}}. We directly finetune GPT-2 and expand its vocabulary to include the item vocabulary.
(7) \underline{\textbf{BART}}. We directly finetune BART and expand its vocabulary to include the same item vocabulary.
(8) \underline{\textbf{DialoGPT}}. We directly finetune DialoGPT and expand its vocabulary to include same item vocabulary.

For our RecInDial, in addition to the full model (9) \underline{\textbf{RecInDial}}, we also evaluate two variants: (10) \underline{\textbf{{RecInDial} \textit{w/o} VP}}, where we remove the vocabulary pointer; and (11) \underline{\textbf{{RecInDial} \textit{w/o} KG}}, where the knowledge graph part is removed.

\paragraph{Evaluation Metrics.}
As we discussed above, the previous works evaluate the recommender and dialogue modules separately. Following the previous setting~\cite{chen-etal-2019-towards,zhou2020improving}, we evaluate the recommender module by  Recall@k (k = 1, 10, 50).
Besides, we also evaluate Recall@k in an end-to-end manner, \textit{i.e.}, to check whether the final produced response contains the target item. In such a setting, the Recall@K score not only depends on whether the ground truth item appears in the top K recommendation list but also reply on if the recommended item is successfully injected into the generated sentences. Therefore, the end-to-end evaluation is fair for all models and applicable for K = 1, 10, 50. 
For the dialogue module, automatic metrics include:
(1) \textbf{Fluency}: perplexity (PPL) measures the confidence of the
generated responses.
(2) \textbf{Relevance}: BLEU-2/4~\cite{papineni2002bleu} and Rouge-L~\cite{lin2004rouge}.
(3) \textbf{Diversity}: Distinct-n (Dist-n)~\cite{li2016diversity} are defined as the number of distinct n-grams divided by the total amount of words. 
Specifically, we use Dist-2/3/4 at the sentence level to evaluate the diversity of generated responses. 
Besides, we also employ Item Ratio introduced in KGSF~\cite{zhou2020improving} to measure the ratio of items in the generated responses.

%% file: tables/dataset_statistic.tex
\begin{table}\small
\centering
{
\begin{tabular}{ll}
\hline
\textbf{Conversations} & \\
\hline
\# of convs & 10006 \\
\# of utterances & 182150 \\
\# of users & 956 \\
avg token length & 6.8\\
avg turn \# & 18.2 \\
\hline
\end{tabular}
}
{
\begin{tabular}{ll}
\hline
\textbf{Movies} & \\
\hline
\# of mentions & 51699 \\
\# of movies & 6924 \\
avg mentions & 7.5\\
max mentions & 1024\\
min mentions& 1\\
\hline
\end{tabular}
}
\vskip -1em
\caption{Statistics of ReDial dataset. ``\#" means number and ``avg" refers to average. }
\label{tab:dataset_statistic}
\vskip -1em
\end{table}

%% file: figures/data_statistic.tex
\begin{figure}[t]
\centering
\subfigure[Movie \# Distribution ]{\label{sfig:statistis_fre}
\includegraphics[width=0.45\linewidth]{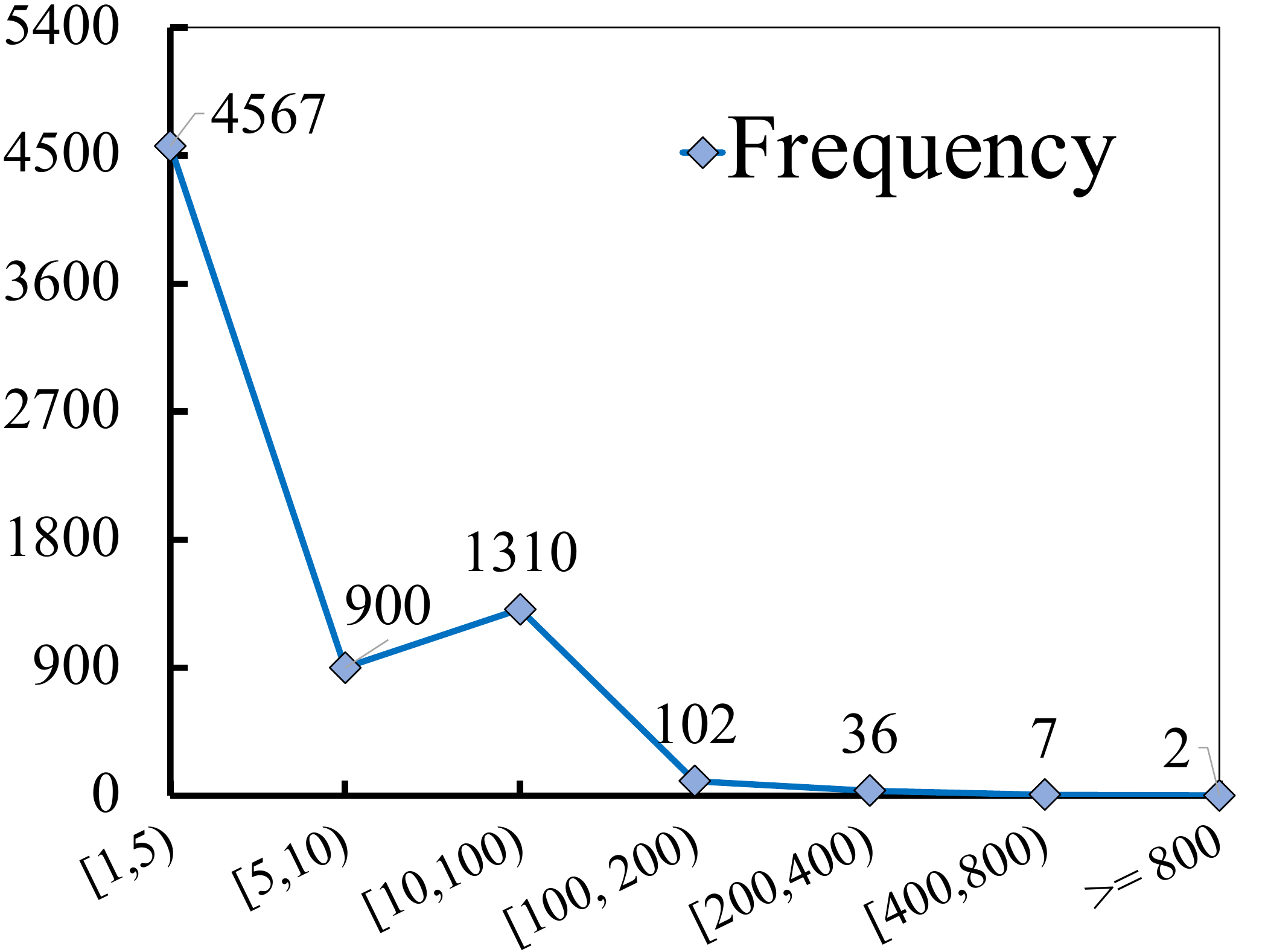}
}
\subfigure[Position Distribution ]{\label{sfig:statistic_turn}
\includegraphics[width=0.45\linewidth]{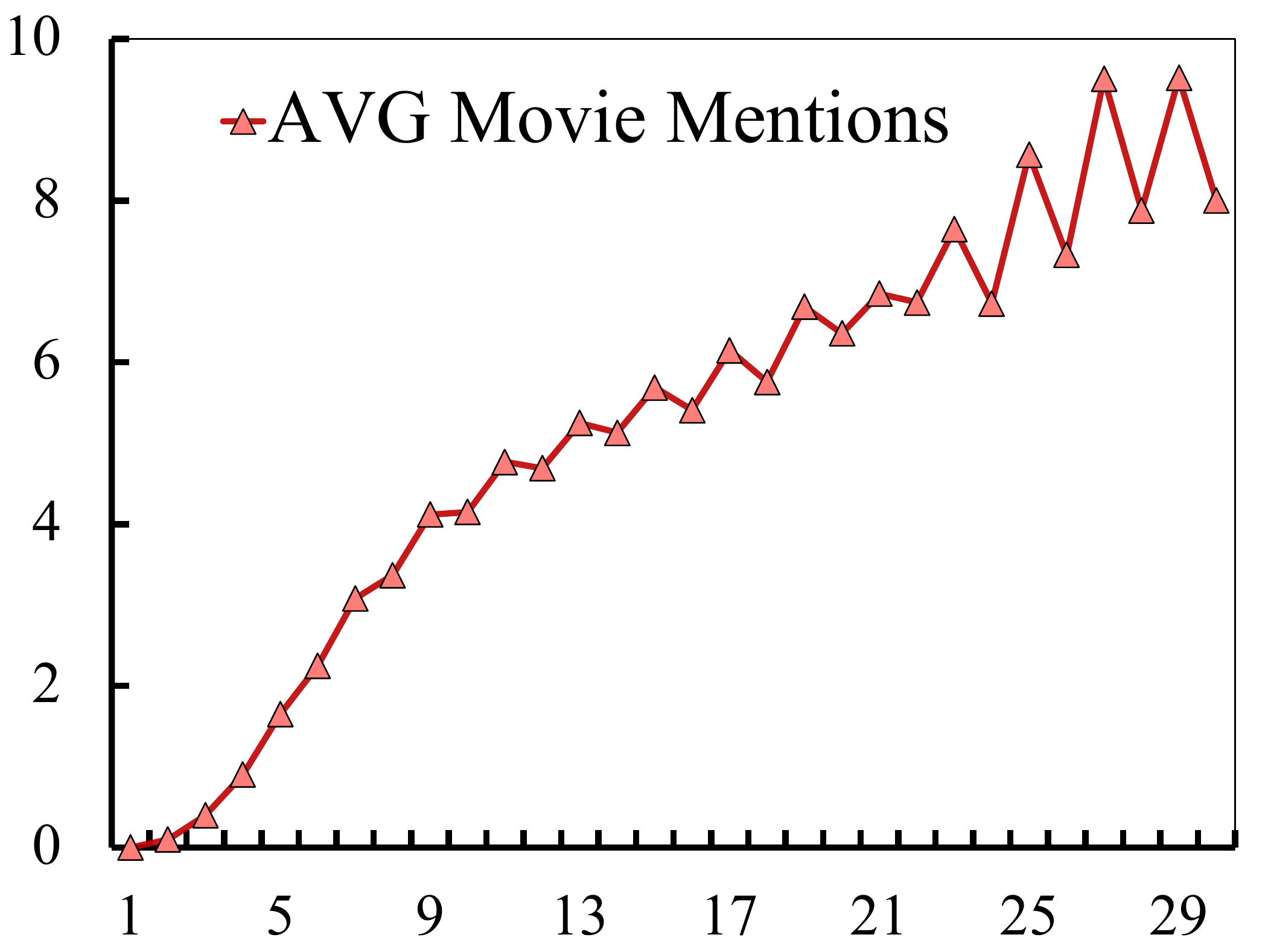}
}
\vskip -1em
\label{fig:statistic}
\caption{For Figure \ref{sfig:statistis_fre}, X-axis: the movie mentions range; Y-axis: movie numbers. For Figure \ref{sfig:statistic_turn}, X-axis: turn positions; Y-axis: average movie mentions.
}
\vskip -1em
\end{figure}

%% file: sections/experimental_results.tex
\section{Experimental Results}
In this section, we first report the comparison results on recommendation and response generation. Then we discuss the human evaluation results. 
After that, we show an example to illustrate how our model works, followed by qualitative analysis.

\subsection{Results on Recommendation}
\label{subsec:results_rec}
The main experimental results for our \textsc{RecInDial} and baseline models on recommendation side are presented in Table~\ref{tab:main_rec}.
And we can draw several observations from the results.

\textit{There is a significant gap between the performance of the recommender module and the performance of the final integrated system.} 
KGSF, the state-of-the-art model, achieves 3.9\% Recall@1 in the recommender module evaluation but yields only 0.9\% in the evaluation of the final produced responses. 
This indicates that the integration strategies utilized by previous methods have significant harm on the recommendation performance.

\textit{Finetuning PLMs on the small CRS dataset is effective.}
As we can see, compared to non-PLM based methods, directly finetuning GPT-2/BART/DialoGPT on the \textsc{ReDial} achieves the obvious performance gain on recommendation.

\textit{Our RecInDial model significantly outperforms the SOTAs on recommendation performance.} 
As shown in Table 2, our RecInDial achieves the best Recall@k (k = 1, 10, 50) scores under the end-to-end evaluation, which demonstrates the superior performance of the PLMs with the unified design.

\subsection{Results on Dialogue Generation}
\label{subsec:results_gen}
\input{tables/main_results_rec}
\input{tables/ablation_results_rec}
\input{tables/main_results_gen}
Since CRS aims to recommend items during natural conversations, we conduct both automatic and human evaluations to investigate the quality of generated responses by RecInDial and baselines.

\cparagraph{Automatic Evaluation.}
\label{subsec:results_automatic}
\indent
Table~\ref{tab:main_gen} shows the main comparison results on Dist-2/3/4, BLEU-2/4, Rouge-L and PPL.
As we can see, RecInDial significantly outperforms all baselines on Dist-n, which indicates that
\textit{PLM helps generate more diverse responses.} 
Previous works suffer from the low-resource issue due to the small crowd-sourcing CRS dataset and tend to generate boring and singular responses. 
On the other hand, \textit{our RecInDial model tends to recommend items more frequently}, 
as the Item Ratio score of RecInDial is much higher than those of baselines.
Besides, our RecInDial and PLM-based methods consistently achieve remarkable improvement over non-PLM based methods on all metrics, which demonstrates the superior performance of PLMs on dialogue generation. 

\cparagraph{Human Evaluation.}
\label{subsec:results_human}
\indent
To further investigate the effectiveness of RecInDial, we conduct a human evaluation experiment, where four crowd-workers are employed to score on 100 context-response pairs that are randomly sampled from the test set. 
Then, we collect the generation results of RecInDial and the baseline models and compare their performance on the following three aspects: 
(1) \textbf{Fluency}. Whether a response is organized in regular English grammar and easy to understand. 
(2) \textbf{Informativeness}. Whether a response is meaningful and not a ``safe response'', and repetitive responses are regarded as uninformative. 
(3) \textbf{Coherence}. Whether a response is coherent with the previous context. 
The crowd-workers give a score on the scale of [0, 1, 2] to show the quality of the responses, and higher scores indicate better qualities. 

\input{tables/human_eval}
We calculate the average score for each model, as well as the ground truth that humans give. 
As shown in Table~\ref{tab:human_eval}, our model shows better performance than all the baselines. Interestingly, ground-truth Human cannot get a 100\% correctness in all the four evaluation metrics. The reason may be that words and phrases sent by human annotators on AMT platform sometimes are the casual usage popular on Internet, which has the wrong grammar. 
For the fluency, all models generate fluent utterances and show similar performance. 
For the informativeness, our RecInDial achieves better performance than the baselines, which indicates RecInDial tends to generate more meaningful responses.

\subsection{Ablation Study}
We then report the performance comparisons on RecInDial's variants. Table \ref{tab:ablation} shows the end-to-end recommendation performance and generation results.
Removing the vocabulary pointer leads to significant drops on R@k and Item Ratio.
\textit{This indicates Vocabulary Pointer (VP) introduced in RecInDial is crucial to the performance of item recommendation.} 
The reason is that the generation process would lose the guidance to switch between general tokens and recommended items without the help of the vocabulary pointer. 
Besides, we can find that \textit{knowledge graph enhanced finetuning helps achieve better recommendation performance.} 
Introducing the node representations learned on the knowledge graph can model the user preference better, which could further enhance the recommendation performance.

\subsection{Qualitative Analysis}
\label{subsec:results_qual}
In this subsection, we present a conversation example to illustrate how our model works in practice. 
\input{figures/case}

In Table~\ref{fig:case}, the \textit{Seeker} states that he likes scary movies. 
Our model successfully captured the keyword of ``scary'' and recommends a famous scary movie ``\textit{It (2017)}'' while the state-of-the-art model \textsc{KGSF} produces a safe response ``Hello!'', which shows our RecInDial can generate the responses that are more coherent with the context. 
Interestingly, after the \textit{Seeker} says he watched the old ``\textit{It (1990)}'', our model recommends another horror movie ``\textit{Psycho (1960)}'' also released in the last century. 
The possible reason is that RecInDial infers the seeker is interested in old horror movies. 
The example in Table~\ref{fig:case} shows that our RecInDial tends to generate a more informative response than KGSF. 
In addition, we find that KGSF always generates ``I would recommend \textit{\textcolor{blue}{Item}}'' (\textit{Item} is replaced with \textit{Get out (2017)} in this example) and ``I would recommend it.''. 
The first response pattern successfully integrates the movie item into the response, while the second fails to make a complete recommendation, which reveals the drawback of the copy mechanism in KGSF.

\subsection{Further Analysis}
\label{subsec:results_further}
\input{figures/recall_analysis}

\paragraph{Analysis on Data Imbalance.}
As we discussed aforementioned, the movie occurrence frequency shows an imbalanced distribution over different movies (see Figure \ref{sfig:statistis_fre}). To investigate the effect, we report the Recall@30 and Recall@50 scores over movie mentioned times in Figure~\ref{sfig:exp_recall_mentions}. 
As we can see, the recall scores for low-frequency movies (with mentioned times less than 10) are much lower than those high-frequency movies (with $>100$ mentions). However, most of the movies (5467 out of 6924 movies) in the \textsc{ReDial} dataset are low-frequency movies, which leads to relatively low results in the overall performance.

\paragraph{Analysis on Cold Start.} 
\textsc{ReDial} dataset suffers from the cold-start problem. 
It is hard for models to recommend precise items in the first few turns of the conversation. 
We report the Recall@30 and Recall@50 scores of our RecInDial over different dialogue turns in Figure~\ref{sfig:exp_recall_turns}. 
Generally, we can see that the recall scores are getting better with richer information gradually obtained from dialogue interactions. 
The scores begin to drop when there are more than 5 turns.
The possible reason is that as the conversation goes deeper, the Seekers are no longer satisfied with the recommended high-frequency movies but prefer more personalized recommendations, which makes it more difficult to predict in practice.

%% file: tables/main_results_rec.tex
\begin{table}[t]\setlength{\tabcolsep}{1.45mm}\small
\newcommand{\tabincell}[2]{\begin{tabular}{@{}#1@{}}#2\end{tabular}}
\begin{center}
\scalebox{0.9}{
\begin{tabular}{lcccccc}
\hline
\multirow{2}{*}{Models} & \multicolumn{3}{c}{ \tabincell{c}{\textbf{Eval on Rec Module}} } & \multicolumn{3}{c}{ \tabincell{c}{\textbf{End-to-End Eval}}}
\\
\cmidrule(lr){2-4}\cmidrule(lr){5-7}
& R@1  & R@10 & R@50 & R@1 &R@10 &R@50 \\
\hline
\underline{\textbf{Baselines}} & & & && & \\
Popularity &1.2 &6.1 &17.9 &1.2 &6.1 &17.9 \\
ReDial &2.4 &14.0 &32.0 &0.7 &4.4 &10.0\\
KBRD &3.1 &15.0 &33.6 &0.8 &3.8 &8.8\\
KGSF &3.9 &18.3 &37.8 &0.9 &4.2 &8.8\\
GPT-2 &-&- &- &1.4&6.5 &14.4 \\
BART &-&- &- &1.5&- &- \\
DialoGPT &-&- &-&1.7 &7.1&13.8\\
\hline

RecInDial &-&- &-& \textbf{3.1} & \textbf{14.0} & \textbf{27.0}\\

\hline
\end{tabular}
}
\end{center}
\vskip -1em
\caption{\label{tab:main_rec} 
Main comparison results on recommendation. R@k refers to Recall@k. RecInDial outperms the baselines significantly ($p$$<$$0.01$, paired t-test).
}
\vskip -0.5em
\end{table}

%% file: tables/ablation_results_rec.tex
\begin{table}[t]\setlength{\tabcolsep}{0.45mm}\small
\newcommand{\tabincell}[2]{\begin{tabular}{@{}#1@{}}#2\end{tabular}}
\begin{center}
\resizebox{\linewidth}{!}{
\begin{tabular}{lccccccc}
\hline
Models& R@1 &R@10 &R@50 &Item Ratio &BLEU &Rouge-L\\

\hline
RecInDial & \textbf{3.1} & \textbf{14.0}&\textbf{27.0} & \textbf{43.5}&\textbf{20.7} & \textbf{17.6}\\
RecInDial \textit{w/o} \textsc{VP}&1.8 &8.8&19.5&17.8&18.5&14.6\\
RecInDial \textit{w/o} \textsc{KG}&2.3 &9.4&20.1&39.8&17.7&12.9 \\

\hline
\end{tabular}
}
\end{center}
\vskip -1em
\caption{\label{tab:ablation} 
Comparison results on ablation study.
}
\vskip -1em
\end{table}

%% file: tables/main_results_gen.tex
\begin{table}[t]\setlength{\tabcolsep}{0.4mm}
\begin{center}
\resizebox{\linewidth}{!}{
\begin{tabular}{lcccccccccc}
\hline 
Models & Dist-2  & Dist-3 & Dist-4& IR & BL-2 & BL-4 & Rouge-L &PPL$\downarrow$ \\
\hline
\underline{\textbf{Baselines}} & & && &&&& \\
Transformer &14.8 &15.1 &13.7 &19.4&-&-&-&- \\
ReDial &22.5 &23.6 &22.8 &15.8 &17.8 &7.4 &16.9&61.7 \\
KBRD & 26.3 &36.8 &42.3 &29.6 &18.5 &7.4 &17.1&58.8 \\
KGSF & 28.9 &43.4 &51.9 &32.5 &16.4 &7.4 &14.3&131.1 \\
GPT-2 &35.4 &48.6 &44.1 &14.5  &17.1 &7.7&11.3&56.3 \\
BART &37.6 &49.0 &43.5 &16.0 &17.8 &9.3&13.1&55.6 \\
DialoGPT &47.6&55.9&48.6&15.9 &16.7 &7.8&12.3&56.0 \\
\hline
RecInDial & \textbf{51.8} & \textbf{62.4} & \textbf{59.8} & \textbf{43.5} &\textbf{20.4} &\textbf{11.0}&\textbf{17.6}&\textbf{54.1} \\

\hline
\end{tabular}
}
\end{center}
\vskip -1em
\caption{\label{tab:main_gen} 
Automatic metrics on generated responses. IR denotes the Item Ratio.
}
\vskip -1em
\end{table}

%% file: tables/human_eval.tex
\begin{table}[t]\setlength{\tabcolsep}{1.8mm}\small
\begin{center}
\scalebox{0.9}{
\begin{tabular}{lcccc}
\hline 
Models & Fluency  & Informative & Coherence & Kappa \\
\hline
HUMAN & 1.93 & 1.70  & 1.69 & 0.80 \\
ReDial & 1.90  &1.28 &1.21 &0.75 \\
KBRD & 1.92 &1.32 &1.26 &0.78\\
KGSF & 1.91 & 1.05 & 1.10 & 0.85 \\
RecInDial & 1.93  & 1.65 & 1.60 & 0.84 \\

\hline
\end{tabular}
}
\end{center}
\vskip -1em
\caption{\label{tab:human_eval} Human evaluation results. 
}
\vskip -1.5em
\end{table}

%% file: figures/case.tex
\begin{table}[ht]
\small
\centering
\begin{tabular}{rp{5.5cm}}
\toprule
\multicolumn{2}{c}{{\bf Conversation Example}} \\
\midrule
\textbf{User}: & I like to watch scary movies \\
\midrule
\textbf{Human}: & Tell me what kind of movie do you like? Did you watch \textit{\textcolor{blue}{It (2017)}}? And \textit{\textcolor{blue}{It (1990)}}? \\
\textbf{KBRD}: & Hi, what kind of movies do you like? \\
\textbf{KGSF}: & Hello! \\
\textbf{RecInDial}: & Have you seen \textit{\textcolor{blue}{It (2017)}}? \\
\midrule
\textbf{User}: & I have seen the old one \\
\midrule
\textbf{Human}: & Did you watch \textit{\textcolor{blue}{The Exorcist (1973)}}? \\
\textbf{KBRD}: & \textit{\textcolor{blue}{Get Out (2017)}} \\
\textbf{KGSF}: & I would recommend \textit{\textcolor{blue}{Get Out (2017)}} \\
\textbf{RecInDial}: & How about \textit{\textcolor{blue}{Psycho (1960)}}? \\
\midrule
\textbf{User}: & I have not seen the new one I do like the exorcist \\
\midrule
\textbf{Human}: &  Ok, you should watch the new one even though I would rather the old one. And did you watch \textit{\textcolor{blue}{The Last Exorcism (2010)}}? I hope I have helped \\
\textbf{KBRD}: & I haven't seen that one \\
\textbf{KGSF}: & I would recommend it \\
\textbf{RecInDial}: & Yes, it was good. I hope you enjoy it. Have a good day. Bye \\
\bottomrule
\end{tabular}
\caption{A conversation example on \textit{\textcolor{blue}{movie item}} recommendation. The responses of \textbf{KBRD}, \textbf{KGSF} and \textbf{RecInDial} are from the test results for corresponding models. \textbf{Human} responses are ground-truth.}
\label{fig:case}
\end{table}

%% file: figures/recall_analysis.tex
\begin{figure}[t]
\centering
\subfigure[Recall over Mentions ]{\label{sfig:exp_recall_mentions}
\includegraphics[width=0.45\linewidth]{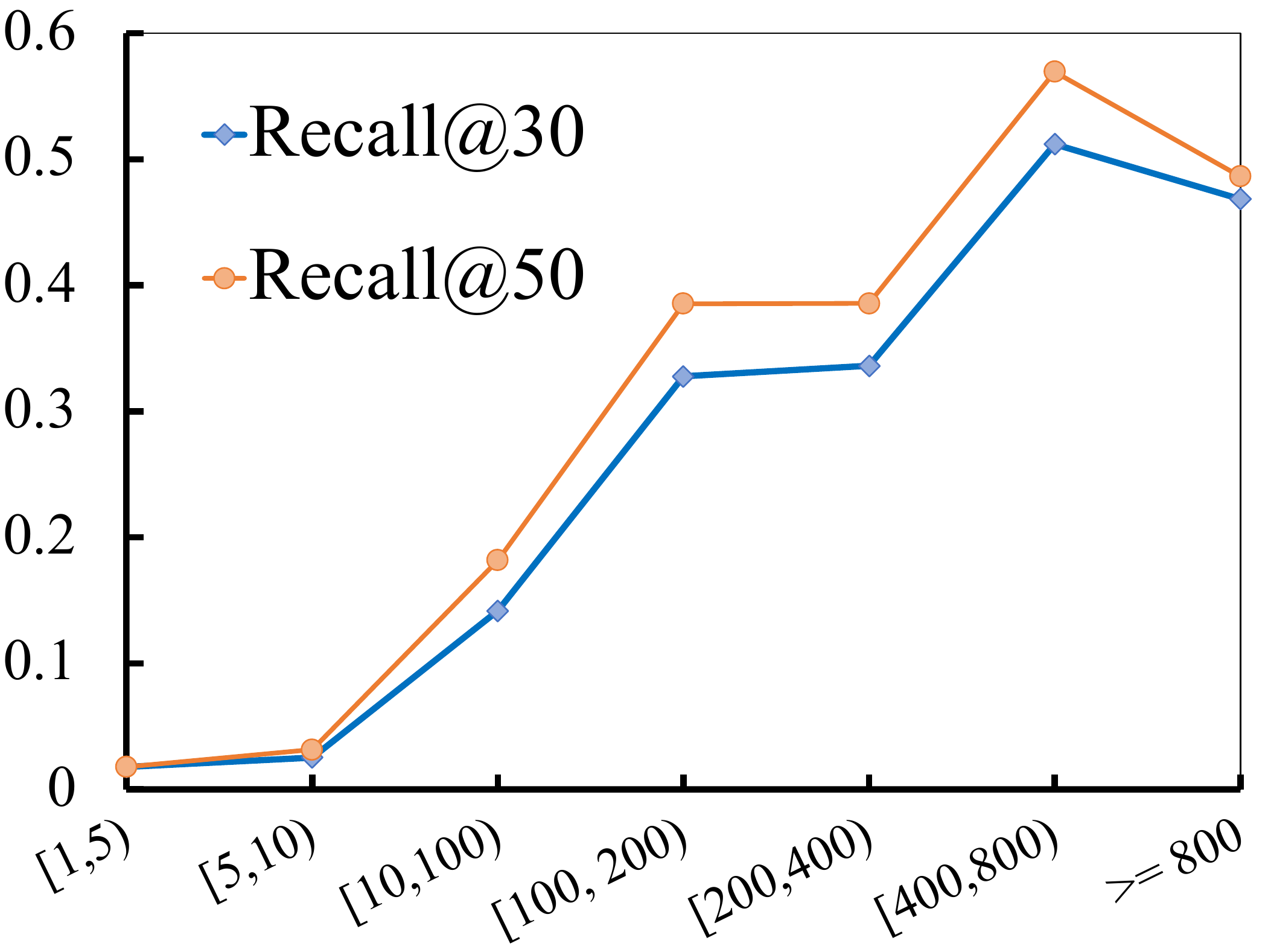}
}
\subfigure[Recall over Turn \#] {\label{sfig:exp_recall_turns}
\includegraphics[width=0.45\linewidth]{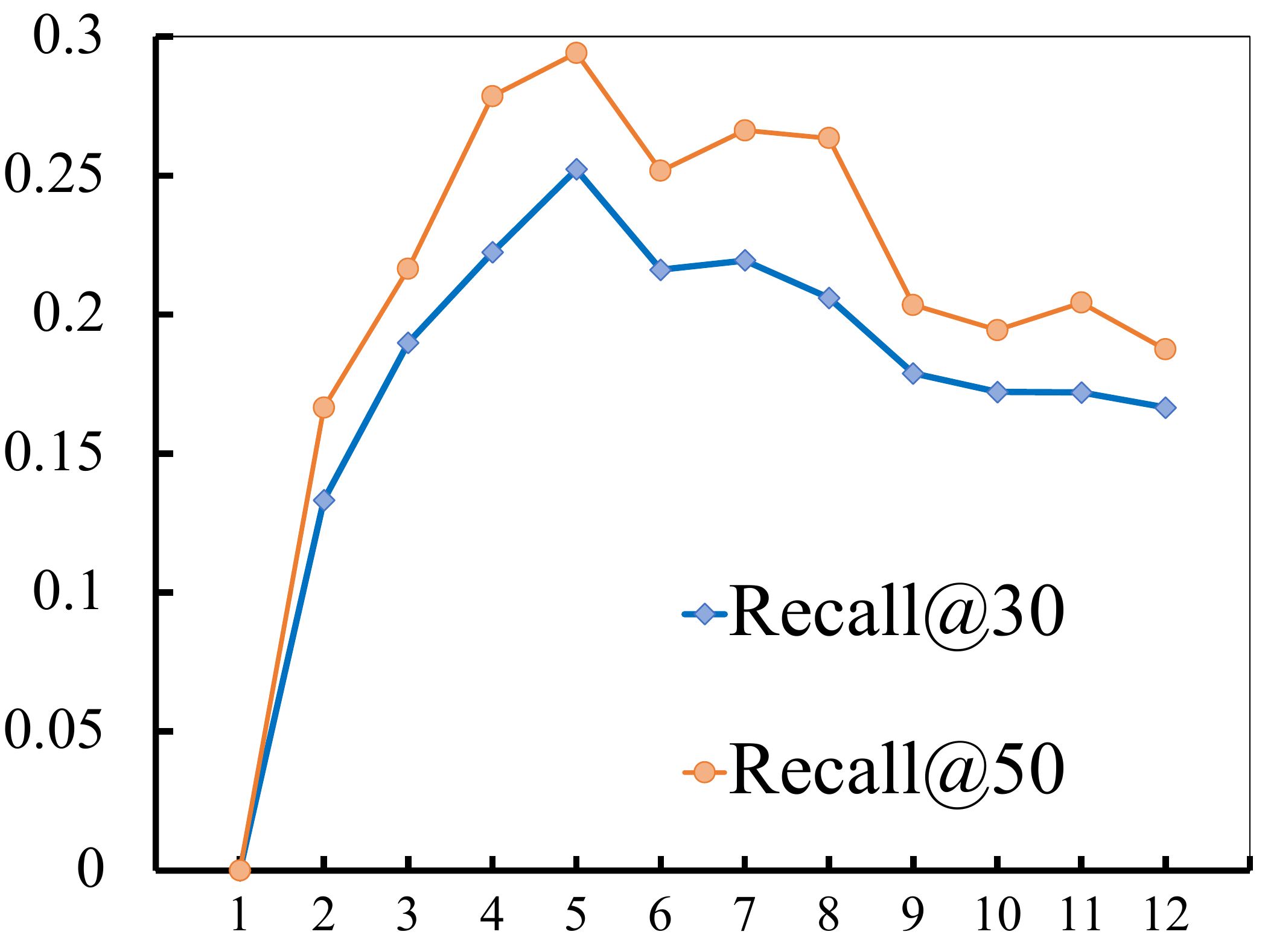}
}
\label{fig:exp_recall_analysis}
\caption{ Y-axis: Recall. For Fig. \ref{sfig:exp_recall_mentions}, X-axis: Movie mentions range. For Fig. \ref{sfig:exp_recall_turns}, X-axis: turn numbers. 
}
\end{figure}

%% file: sections/conclusion.tex
\section{Conclusion}
This paper presents a novel unified PLM-based framework called \textit{RecInDial} for CRS, which integrates the item recommendation into the generation process. 
Specifically, we finetune the large-scale PLMs together with a relational graph convolutional network on an item-oriented knowledge graph. Besides, we design a vocabulary pointer mechanism to unify the response generation and item recommendation into the existing PLMs. 
Extensive experiments on the CRS benchmark dataset \textsc{ReDial} show that RecInDial significantly outperforms the state-of-the-art methods.

\section*{Acknowledgements}
We would like to thank the anonymous reviewers for their feedback and suggestions. The research described in this paper is partially supported by HKSAR ITF No. ITT/018/22LP. 